\documentclass[sigconf,natbib=false]{acmart}

\AtBeginDocument{%
  }

\copyrightyear{2024}
\acmYear{2024}
\setcopyright{acmlicensed}\acmConference[HRI '24 Companion]{Companion of the 2024 ACM/IEEE International Conference on Human-Robot Interaction}{March 11--14, 2024}{Boulder, CO, USA}
\acmBooktitle{Companion of the 2024 ACM/IEEE International Conference on Human-Robot Interaction (HRI '24 Companion), March 11--14, 2024, Boulder, CO, USA}
\acmDOI{10.1145/3610978.3640738}
\acmISBN{979-8-4007-0323-2/24/03}

\usepackage{amsmath,amsfonts}

\usepackage[utf8]{inputenc}
\usepackage{multirow}
\usepackage{csquotes}
\usepackage{array}
\usepackage{adjustbox}
\usepackage{booktabs}
\usepackage{tabularray}
\usepackage{subcaption}
\usepackage[font=small]{caption}
\usepackage{xcolor}
\usepackage{url}
\usepackage{graphicx}
\usepackage{collectbox}
\usepackage{multirow}
\usepackage{makecell}
\usepackage{algorithm}
\usepackage{float}
\makeatletter
\let\NAT@parse\undefined
\makeatother
\usepackage{hyperref}
\usepackage{algpseudocode}

\RequirePackage[
  datamodel=acmdatamodel,
  style=acmnumeric,
  ]{biblatex}

\begin{document}

%%
%% The "title" command has an optional parameter,
%% allowing the author to define a "short title" to be used in page headers.
\title{Transition State Clustering for\\Interaction Segmentation and Learning}
\renewcommand{\shorttitle}{Transition State Clustering for Interaction Segmentation and Learning}
%%
%% The "author" command and its associated commands are used to define
%% the authors and their affiliations.
%% Of note is the shared affiliation of the first two authors, and the
%% "authornote" and "authornotemark" commands
%% used to denote shared contribution to the research.
% \author{Submission \#1237}

% \author{Vignesh Prasad}
% \email{ vignesh.prasad@tu-darmstadt.de}
% \affiliation{%
%   \institution{TU Darmstadt}
%   \city{Darmstadt}
%   \country{Germany}
% }
\author{Fabian Hahne}
\affiliation{%
  \institution{Technische Universität Darmstadt}
  \city{Darmstadt}
  \country{Germany}
}
\email{fabian.hahne@stud.tu-darmstadt.de}
\author{Vignesh Prasad}
\affiliation{
   \institution{Technische Universität Darmstadt}
   \city{Darmstadt}
   % \state{Hesse}
   \country{Germany}
 }
\email{vignesh.prasad@tu-darmstadt.de}

 \author{Alap Kshirsagar}
 \affiliation{
   \institution{Technische Universität Darmstadt}
   \city{Darmstadt}
   % \state{Hesse}
   \country{Germany}
 }
\email{alap.kshirsagar92@gmail.com}

\author{Dorothea Koert}
\affiliation{
   \institution{Technische Universität Darmstadt}
   \city{Darmstadt}
   % \state{Hesse}
   \country{Germany}
 }
\email{dorothea.koert@tu-darmstadt.de}

\author{Ruth Maria Stock-Homburg}
\affiliation{
   \institution{Technische Universität Darmstadt}
   \city{Darmstadt}
   % \state{Hesse}
   \country{Germany}
 }
 \email{rsh@bwl.tu-darmstadt.de}
\author{Jan Peters}
\affiliation{
   \institution{Technische Universität Darmstadt}
   \city{Darmstadt}
   % \state{Hesse}
   \country{Germany}
 }
 \affiliation{
   \institution{German Research Center for AI}
   \city{Darmstadt}
   % \state{Hesse}
   \country{Germany}
 }
  \affiliation{
   \institution{Hessian Center for AI}
   \city{Darmstadt}
   % \state{Hesse}
   \country{Germany}
 }
\email{mail@jan-peters.net}

\author{Georgia Chalvatzaki}
\affiliation{
   \institution{Technische Universität Darmstadt}
   \city{Darmstadt}
   % \state{Hesse}
   \country{Germany}
 }
\affiliation{
   \institution{Hessian Center for AI}
   \city{Darmstadt}
   % \state{Hesse}
   \country{Germany}
 }
% \affiliation{
%    \institution{Center for Mind, Brain \& Behavior}
%    \city{Uni. Marburg \& JLU Giessen}
%    % \state{Hesse}
%    \country{Germany}
% }
\email{georgia.chalvatzaki@tu-darmstadt.de}

%%
%% By default, the full list of authors will be used in the page
%% headers. Often, this list is too long, and will overlap
%% other information printed in the page headers. This command allows
%% the author to define a more concise list
%% of authors' names for this purpose.
\renewcommand{\shortauthors}{Fabian Hahne et al.}
%% No italics

%%
%% The abstract is a short summary of the work to be presented in the
%% article.
\begin{abstract}
Hidden Markov Models with an underlying Mixture of Gaussian structure have proven effective in learning Human-Robot Interactions from demonstrations for various interactive tasks via Gaussian Mixture Regression. However, a mismatch occurs when segmenting the interaction using only the observed state of the human compared to the joint state of the human and the robot. To enhance this underlying segmentation and subsequently the predictive abilities of such Gaussian Mixture-based approaches, we take a hierarchical approach by learning an additional mixture distribution on the states at the transition boundary. This helps prevent misclassifications that usually occur in such states. We find that our framework improves the performance of the underlying Gaussian Mixture-based approach, which we evaluate on various interactive tasks such as handshaking and fistbumps.
\end{abstract}

%%
%% The code below is generated by the tool at http://dl.acm.org/ccs.cfm.
%% Please copy and paste the code instead of the example below.
%%
\begin{CCSXML}
<ccs2012>
   <concept>
       <concept_id>10003752.10010070.10010071.10010074</concept_id>
       <concept_desc>Theory of computation~Unsupervised learning and clustering</concept_desc>
       <concept_significance>500</concept_significance>
       </concept>
   <concept>
       <concept_id>10010147.10010257.10010258.10010260.10010267</concept_id>
       <concept_desc>Computing methodologies~Mixture modeling</concept_desc>
       <concept_significance>300</concept_significance>
       </concept>
   <concept>
       <concept_id>10010147.10010257.10010282.10010290</concept_id>
       <concept_desc>Computing methodologies~Learning from demonstrations</concept_desc>
       <concept_significance>100</concept_significance>
       </concept>
 </ccs2012>
\end{CCSXML}

\ccsdesc[500]{Theory of computation~Unsupervised learning and clustering}
\ccsdesc[300]{Computing methodologies~Mixture modeling}
\ccsdesc[100]{Computing methodologies~Learning from demonstrations}

%%
%% Keywords. The author(s) should pick words that accurately describe
%% the work being presented. Separate the keywords with commas.
\keywords{Hidden Markov Models, Learning from Demonstrations}

% \received{20 February 2007}
% \received[revised]{12 March 2009}
% \received[accepted]{5 June 2009}

%%
%% This command processes the author and affiliation and title
%% information and builds the first part of the formatted document.
\maketitle

\section{Introduction}

Human-Human Interactions involve various non-verbal gestures, like handshakes, fostering societal trust and affiliation~\cite{phutela2015importance}. In the context of Human-Robot Interactions (HRI), non-verbal communication is crucial for robot acceptance, requiring robots to comprehend and execute social actions seamlessly without appearing uncanny to the human participants in the interaction~\cite{mori2012uncanny}. Learning these interactions is challenging due to human diversity and subtle variations in actions. To address this problem of predicting accurate response behaviors in Human-Robot Interactions, Learning from Demonstrations (LfD) approaches have shown good performance by learning joint distributions over the observations of the human and the robot~\cite{lee2017survey}.

Since many interactive tasks can naturally be broken down into underlying segments or phases that are then sequenced to achieve suitable behavior, previous works have explored learning HRI from demonstrations using Gaussian Mixture Models (GMMs) or, additionally, Hidden Markov Models (HMMs) with an underlying Mixture of Gaussians structure~\cite{calinon2009learning,evrard2009teaching,rozo2016learning1,rozo2016learning2,10000239,PIGNAT201761,vogt2017system}. However, Hidden Markov Models have limitations in representing transition states. In complex Human-Robot-Interaction settings, where dynamics are intricate and context-dependent, the simplistic representation of transitions may not capture the nuanced nature of human behaviors.

\begin{figure}
    \centering
    \includegraphics[width=0.9\linewidth]{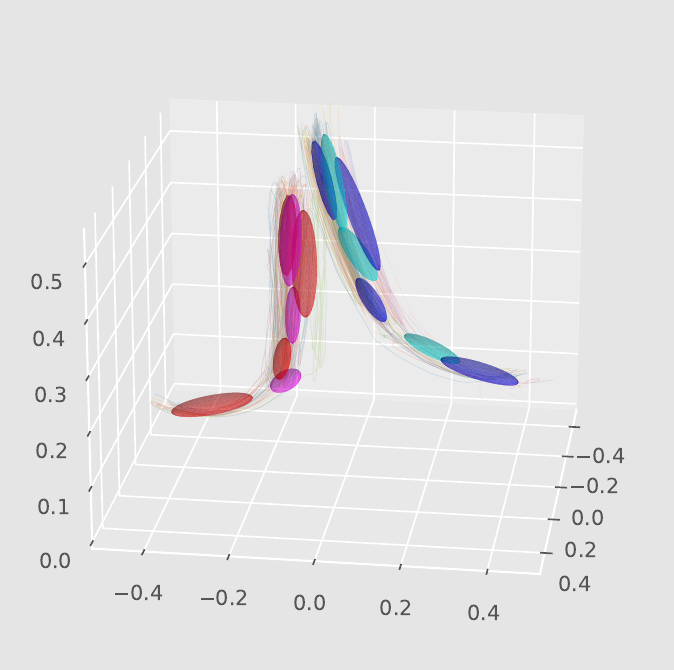}
    \caption{An example of the hidden states (Human - red, Robot - blue) learned by an HMM and the learned transition state clusters (Human - magenta, Robot - cyan) from demonstrations of handshaking.}
    \label{fig:teaser}
\end{figure}

\begin{figure*}
    \centering
    \includegraphics[width=\textwidth]{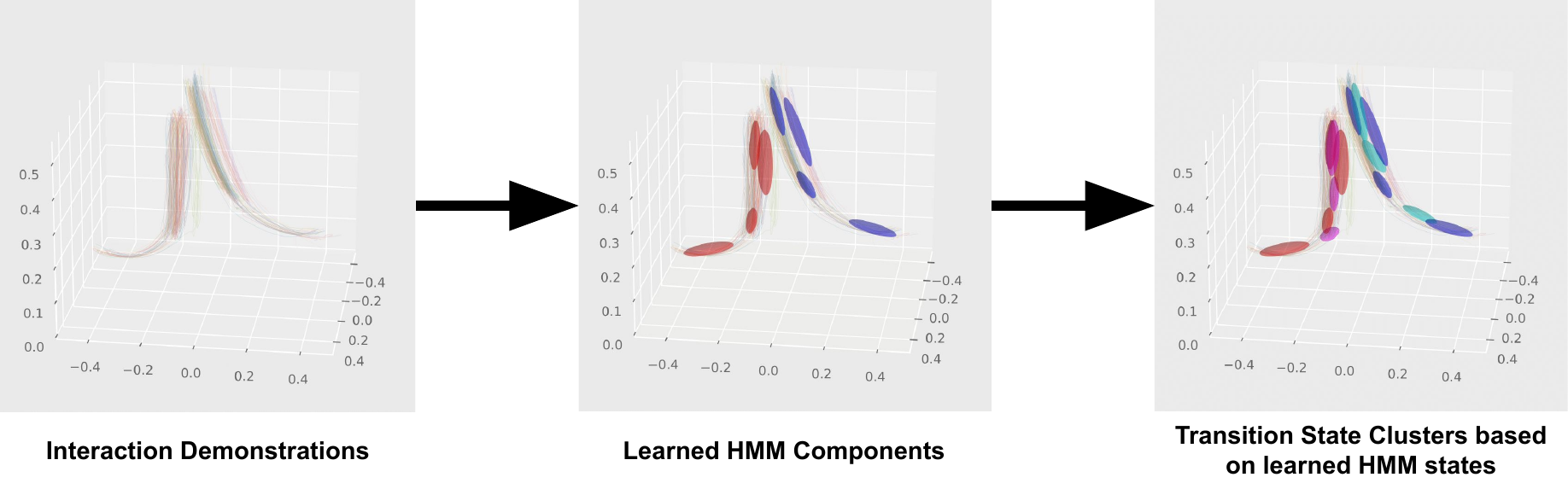}
    \caption{An overview of our proposed approach. Given demonstrations of an interaction (left), such as end-effector trajectories when performing a handshake, we first learn an HMM over the demonstrations (middle) to segment the interaction into underlying phases (red - human, blue - robot). Based on the learned HMM, we subsequently learn an additional distribution over the observations near the transition boundaries of the HMM hidden states, as shown in the image on the right (magenta - human, cyan - robot).}
    \label{fig:overview}
\end{figure*}

To enhance the segmentation abilities of GMM/HMM-based approaches, incorporating a separate mixture distribution over observations at the transition boundary between two underlying Gaussian states has demonstrated effectiveness in identifying key change points in a trajectory~\cite{krishnan2017transition}. This leads to a hierarchical model with the first level learning from demonstrated trajectories and the second level focusing on \enquote{transition states} (Fig.~\ref{fig:teaser}). In this paper, we investigate how this hybrid model concept can enhance the performance of GMM/HMM approaches in learning Human-Robot Interaction (HRI). We demonstrate that the additional Transition State Clustering, implemented on top of an HMM model trained over the states of both the human and the robot, improves segmentation when using only human observations. Consequently, this enhancement leads to improved conditional predictions of robot actions based on human observations.

% our Human-Robot-Interaction (HRI) approach  from motion-captured Human-Human Interaction (HHI) data. Using a Hidden Markov Model (HMM) with a Multivariate Normal Distribution (MVN) as underlying distribution, we initially learn the joint distributions to segment trajectories. Subsequently, a second Hidden Markov Model over transition states improves transition modeling, allowing the robot to better generalize and adapt.

\section{Preliminaries}
In this section, we provide a short overview of the fundamental concepts used in our approach, namely Hidden Markov Models (Sec.~\ref{HMM-foundation}) and Transition State Clustering (Sec.~\ref{TSC-foundation}). 
% \subsection{Imitation Learning} \label{imitation}
% Imitation Learning, also referred to as Programming by Demonstration or Learning from Demonstration, is a prevalent machine learning technique in robotics and serves as the foundation for our proposed approach~\cite{Billard2008}. This method replicates how humans and various animals naturally acquire new skills, aiming to mimic the actions of skilled instructors in specific tasks~\cite{ARGALL2009469, 10.1145/3054912}. Widely applied in robotics, particularly in human-robot interaction, imitation learning eliminates the need for explicit programming or designing reward functions, as is typically necessary in a reinforcement learning approach. Therefore, it empowers individuals with limited programming or robotics expertise to effectively instruct agents in task execution~\cite{10.1145/3054912}.
\subsection{Hidden Markov Models} \label{HMM-foundation}
Hidden Markov Models (HMMs) are probabilistic models derived from Discrete Markov Models and represent a sequence of observations via underlying hidden states. In HMMs, states are not directly observable but can only be inferred through another set of stochastic processes that generate the sequence of observations. Generally, an HMM represents a sequence of observations $\boldsymbol{o}_{1:T}$ as a sequence of $S$ hidden latent states with some emission probability. The HMM is characterized by its initial state distribution $\pi_i$ over the states $i \in \{ 1 \dots S \}$ and the transition probability $\mathcal{T}_{i,j}$, describing the probability of transitioning from state $i$ to state $j$. In our case, the emission probabilities for each state  $i \in \{ 1 \dots S \}$ are characterized with a normal distribution  $\mathcal{N}(\boldsymbol o_t; \boldsymbol\mu_i,\boldsymbol\Sigma_i)$ with mean $\boldsymbol \mu_i$ and covariance $\boldsymbol \Sigma_i$. This is, in essence, similar to learning a Gaussian Mixture Model over the observations followed by learning the transitions between the different components. Further information about Hidden Markov Models can be found in~\cite{18626}.
\par
Once the underlying states are learned, HMMs predict the belief of the hidden states based on the observations via the learned forward variable $h_i(o_t)$ that denotes the hidden state probability of the current observation based on the history
\begin{equation} \label{eq:normalized_forward}
    h_i(\boldsymbol{o}_t) = \frac{\alpha_i(\boldsymbol{o}_t)}{\sum\nolimits_{k=1}^{S}\alpha_k(\boldsymbol{o}_t)}
\end{equation}
where
\begin{equation} \label{eq:forward}
  \alpha_i(\boldsymbol{o}_t) = \mathcal{N}(\boldsymbol{o}_t;\boldsymbol{\mu}_i, \boldsymbol{\Sigma}_i)\sum_{k=1}^S\alpha_k(\boldsymbol{o}_{t-1})\mathcal{T}_{k,i}
\end{equation}
and $\alpha_i(\boldsymbol{o}_0) = \pi_i$. The forward variable describes the probability of observing the sequence $\boldsymbol o_{1:t}$ and ending up in state $i$ at time step $t$.
\par
The HMMs are trained using the Baum-Welch algorithm which is a special case of the Expectation-Maximization (EM) algorithm. The goal is to learn the model parameters $\boldsymbol \mu_i$ and $\boldsymbol \Sigma_i$ such that it maximizes the probability of observing a given sequence, $\boldsymbol o_{1:t}$. For more information on learning Hidden Markov Models in the context of robot learning, we refer to~\cite{calinon2016tutorial,PIGNAT201761}
\par
To learn a joint distribution between both interacting partners, we concatenate Degrees of Freedom (DoFs) of both agents, resulting in the following decomposition of the learned distributions
\begin{equation} \label{eq:mean}
    \boldsymbol\mu_i = \begin{bmatrix}
        \boldsymbol\mu_i^1 \\
        \boldsymbol\mu_i^2
    \end{bmatrix}
\end{equation}
and
\begin{equation} \label{eq:cov}
    \boldsymbol\Sigma_i = \begin{bmatrix}
        \boldsymbol\Sigma_i^{11} & \boldsymbol\Sigma_i^{12}\\
        \boldsymbol\Sigma_i^{21} & \boldsymbol\Sigma_i^{22}
    \end{bmatrix}
\end{equation}
for the mean and covariance. This decomposition can then be used to conditionally predict the robot's actions $\boldsymbol o_{1:t}^2$ from the observed human motions $\boldsymbol o_{1:t}^1$ as
\begin{equation} \label{eq:reconstruction}
    \boldsymbol o_{1:t}^2 = \sum_{i = 1}^{S} \frac{\alpha_i(\boldsymbol o_{1:t}^1)}{\sum_{k = 1}^{S}\alpha_k(\boldsymbol o_{1:t}^1)}(\boldsymbol\mu_i^1 + \boldsymbol\Sigma_i^{21}(\boldsymbol\Sigma_i^{12})^{-1}(\boldsymbol\mu_i^1- \boldsymbol o_{1:t}^1))
\end{equation}
where $\alpha_k(\boldsymbol o_{1:t}^1)$ is the forward variable calculated using the marginal distribution for the human DoFs.

\subsection{Transition State Clustering} \label{TSC-foundation}
Transition State Clustering (TSC) is an unsupervised segmentation algorithm proposed by Krishnan et. al.~\cite{krishnan2017transition}. 
% Given a set of demonstrations D, where each demonstration $d_i \in D$ is a trajectory of length N. Thus $d_i = [x_1,x_2,...,x_N]$. These demonstrations are modeled as a noisy dynamical system with dynamics $\zeta$ and Gaussian white noise w such that:
% $$x_{t+1} = \zeta(x_t) + w_t.$$
% Assuming local linearity and modeling this as a switched linear system, we have:
% $$x_{t+1} = A_tx_t + w_t$$
% for one of m $d\times d$ matrices $\{A^{(1)},A^{(2)},...,A^{(m)}\}$.
% where $A_t$ can be one of m $d\times d$ matrices $\{A^{(1)},A^{(2)},...,A^{(m)}\}$
% \par
% Any state where $A_t \neq A_{t+1}$ is called a transition state. The set of all transition states, denoted as $\Gamma$, over all demonstrations D can be modeled as an underlying parameterized function over state space x and time t:
% $$\Gamma \sim f_{\theta}(x,t)$$
Given a set of demonstrations, the goal of Transition State Clustering is to fit mixture models to the demonstrations.
To identify the transition states, initially a Gaussian Mixture Model (GMM) is fitted to the demonstrated trajectories and each observation $\boldsymbol{o}_t$ is assigned to its most likely mixture component $c_t$. Subsequently, every state where $c_t \neq c_{t-1}$ is marked as a transition state after which a second GMM is fitted over the transition state, thus grouping events that trigger the transitions between the first GMM. For a more in-depth understanding of Transition State Clustering, we refer the reader to~\cite{krishnan2017transition}.

\section{Interaction Segmentation and Learning}
In this section, we introduce our proposed approach, which uses Transition State Clustering (TSC) with Hidden Markov Models (HMMs) for learning Human-Robot Interaction. We perform trajectory segmentation with TSC  by using an HMM as the underlying distribution and subsequently using the forward variable of the HMM to identify transition states. We then use the proposed HMM-TSC model for conditionally generating the robot's actions based on the human's observations instead of just using the model for segmentation as in~\cite{krishnan2017transition}.
% We describe the training if the HMM in Sec.~\ref{hmm-init}, followed by our adaption of the TSC algorithm in Sec.~\ref{tsc-init}. Finally, we describe, how a trajectory can be reconstructed from the trained TSC-HMM in Sec.~\ref{reconstruction}.
% \subsection{Hidden Markov Model Initialization and Training} \label{hmm-init}
% First, before training the HMM, it has to be initialized. The initialization occurs with a manually defined number of states n as well as an arbitrary amount of training data. The training data consists of the concatenated Cartesian joint positions and velocities of both actors.
% \newline
% Initially, the training data is split into n segments of equal size along time t, where each segment corresponds to the model's states. Each segment and its observations are described by a Multivariate Normal Distribution. Then, to train the model the Baum-Welch alogrithm is used.

% \subsection{Transition State Clustering} \label{tsc-init}
% To improve accuracy, a second model is trained on the first model's transition states. These transition states are defined similarly to Krishnan et. al.~\cite{krishnan2017transition} as described in Section~\ref{TSC-foundation}.
% \newline
% Instead of modeling the states as a switched linear system and using the dynamics matrix $A_t$ to identify the transition states, we make use of the Hidden Markov Model's forward variable $h_i$. 
Instead of only considering the state right before a transition as a transition state, we consider an arbitrary number of states around the transition as transition states (Fig.~\ref{fig:overview}). %Furthermore, we only consider the time-independent variant and thus do not train an additional model along the time axis, but instead use positions as well as velocities in the input demonstrations.

\begin{figure}
    \centering
    \includegraphics[trim={10 140 0 25},clip,width=\linewidth]{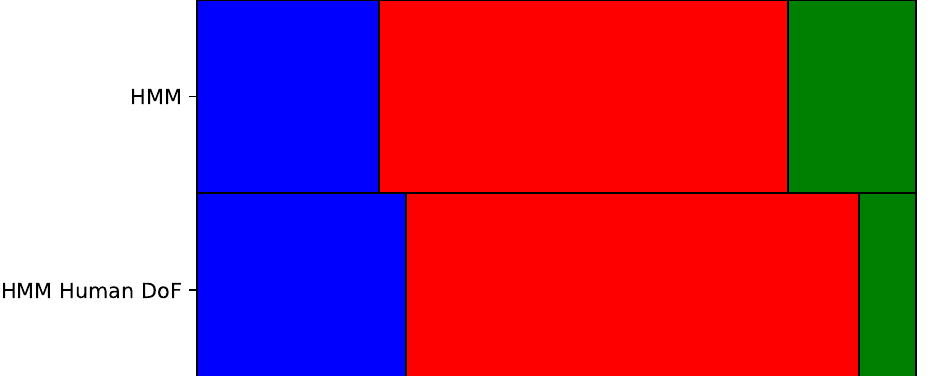}
    \caption{Example of the predicted segments when using the combined DoFs of the human and the robot (top row) and only the human DoFs (bottom row).}
    \label{fig:segment-mismatch}
\end{figure}
Moreover, we train the HMM with the combined observations of the human and the robot, however, during test time, we use only the human observations to calculate the HMM forward variable and subsequently condition the HMM to predict the robot motions. %Since we train on the combined observations but use only the human observations during test time, %a mismatch exists between the forward variable prediction in the two cases, which we find occurs at states near the transition boundary. Therefore, we learn an additional HMM over the observations in 
By doing so, we find that a mismatch exists between the HMM forward variable calculated only with the human's observations $h_{i}(o_{t}^1)$, and the HMM forward variable calculated with the observations of both the human and the robot $h_{i}(o_{t}^{1,2})$. We found that these observations where $h_{i}(o_{t}^1) \neq h_{i}(o_{t}^{1,2})$ usually occurred around transition boundaries between the hidden states (Fig.~\ref{fig:segment-mismatch}). Therefore, we learn a second HMM over the set of all observations $o_{t}$ where $h_{i}(o_{t}^1) \neq h_{i}(o_{t}^{1,2})$ which give us the Transition state Clusters. The approach is outlined in algorithm \ref{alg:tsc}. We then use the resulting TSC-HMM model for conditionally generating the trajectory of the robot given the observations of the human as shown in Eq.~\ref{eq:reconstruction}.
\begin{algorithm}
    \caption{Transition State Clustering Adaption}
    \label{alg:tsc}
    \begin{algorithmic}
        \State \textbf{Input:}
        \State \hspace{5mm} 1. training data $\boldsymbol{o}_{1:T}$
        \State \hspace{5mm} 2. number of states $n$
        \State \hspace{5mm} 3. trained HMM $\lambda$
        \State \textbf{Output:}
        \State \hspace{5mm} 1. trained HMM $\lambda_T$
    \end{algorithmic}
    \vspace{3mm}
    \begin{algorithmic}[1]
        \State Identify transition states wherever $h_{i}(o_{t}^1) \neq h_{i}(o_{t}^{12})$ of $\lambda$
        \State Initialize a new HMM $\lambda_T$ using transition states as training data
        \State Train $\lambda_T$ according to the Baum-Welch EM algorithm
        \State \Return $\lambda_T$
    \end{algorithmic}
\end{algorithm}

\begin{figure*}
    \centering
    \includegraphics[width=0.3\textwidth]{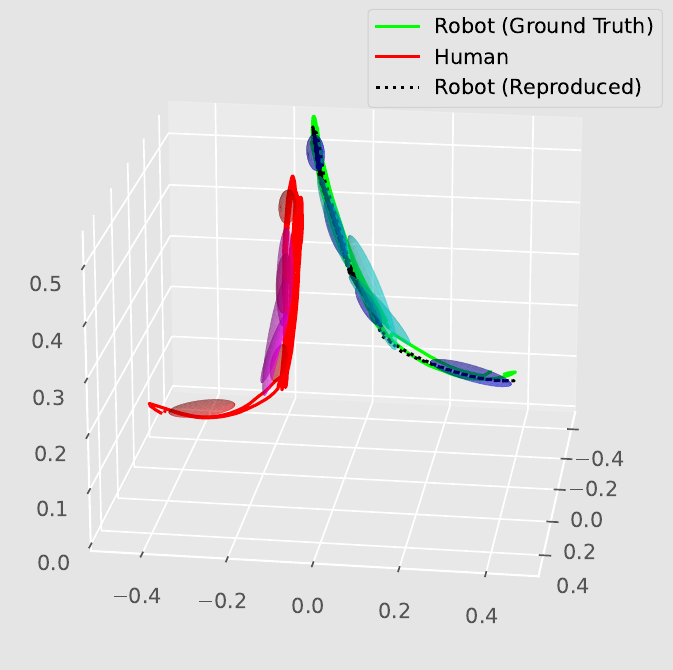}\hfill
    \includegraphics[width=0.3\textwidth]{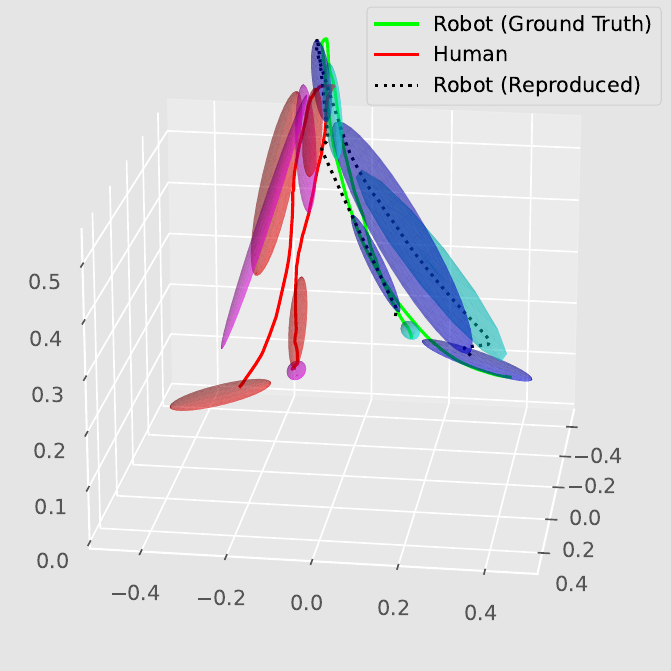}\hfill
    \includegraphics[width=0.3\textwidth]{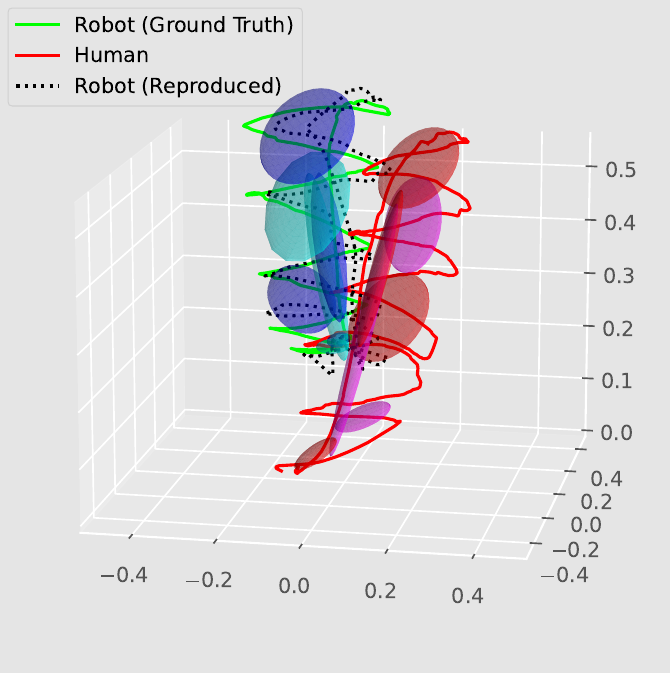}
    \caption{This figure shows example 3D plots of reconstructed trajectories for the different interactions considered in the work. Each plot consists of the input trajectory of the human, the ground truth trajectory for the robot and the reconstructed trajectory for the robot, along with the Gaussian states of the HMM and the transition state clusters.}
    \label{fig:results}
\end{figure*}

\section{Experiments and Results}
% In this section we first discuss the datasets we used (Sec.~\ref{datasets}) as well as the our experimental setup (Sec.~\ref{setup}). Then we present the results of predicting the controlled agent’s trajectories (Sec.~\ref{results})
% \subsection{Datasets}\label{datasets}
We use a dataset of social interactions~\cite{butepage2020imitating} which
consists of 3D Cartesian coordinates of two human partners interacting with one another captured using Rokoko motion capture suits at a frequency of 40Hz.  From the dataset in~\cite{butepage2020imitating}, we use the handshake interactions 
and two types of fistbumps. The first, \enquote{rocket fistbump}, involves partners bumping their fists at a waist level and raising them vertically to a suitable height and returning to a neutral pose. The second is called \enquote{parachute fistbump} where partners bump their fists at a shoulder height and then oscillate them horizontally while moving downwards. % An example of the handshake interaction can be found in Figure \ref{fig:dataset}.
% The second dataset we use is a bimanual handover dataset~\cite{kshirsagar2023dataset} which consists of 3D cartesian joint coordinates of bimanual (dual arm) human-human handover recordings from 12 participants with 10 distinct objects. The trajectories were captured using an OptiTrack motion tracking system at a frequency of 120Hz. 
We use the hand trajectories of the first partner as the Human DoFs and the hand trajectories of the second partner as the robot DoFs. Along with the 3D positions, we use the position differences between the timesteps as well, which acts as a proxy for the velocity. %For the handovers, we consider the Robot-to-Human handover scenario where the robot is the giver and the human is the receiver. %Both datasets were split into test and training datasets. The training dataset consists of 15 trajectories of the respective interactions, with the leftover trajectories representing the test datasets.
% \begin{figure}
%     \centering
%     \includegraphics[trim={10 140 0 25},clip,width=0.8\linewidth]{images/buetepage.png}
%     \caption{Example of the handshake interaction recorded by Bütepage et. al.~\cite{butepage2020imitating}}
%     \label{fig:dataset}
% \end{figure}
% \subsection{Experimental Setup}\label{setup}
The approach is implemented using the Python version of PbDlib developed by Pignat et al.~\cite{PIGNAT201761}. The Baum-Welch algorithm for training the Hidden Markov Models runs for a maximum of 40 iterations. If the change in the log-likelihood does not exceed $10^{-4}$, the algorithm is considered converged, and the Baum-Welch algorithm halts. The HMM and the TSC model are initialized by temporally dividing the corresponding input trajectories into $S$ bins and calculating each bin's mean and covariance. A regularization factor is added to the diagonal elements of the covariance matrices to prevent numeric instabilities. We used the same regularization factor of $10^{-2}$ for both the initial HMM and the TSC-HMM. For the dataset in~\cite{butepage2020imitating}, we used 4 states for the initial HMM and 3 states for the TSC-HMM. %For the handover dataset~\cite{kshirsagar2023dataset} we used 3 states for the initial HMM and 2 states for the TSC-HMM. 
These parameters were determined using empirical testing.
We sample a random batch of 15 training demonstrations as the input data for training the model. We run the training with 100 different random seeds and consequently, 100 different random batches of training samples thus providing a broad distribution of samples of each one of the interactions. To gauge the performance of the model, the Mean Squared Error between the predicted trajectory and the ground truth trajectory of the robot DoFs is used as a metric.

% \subsection{Conditioned Prediction Results}\label{results}
% To get an accurate performance assessment, we compare our approach to two other approaches. As a baseline comparison for our approach, we used a simple Hidden Markov Model. Additionally, as a second comparison, we used a modified TSC-HMM. However, instead of initializing the second HMM with the set of transition states, we use the trained initial Hidden Markov Model as a 'warm start'. Thus we copy the initial Hidden Markov Model's mean $\boldsymbol \mu_i$ and covariance $\boldsymbol \Sigma_i$ to the second Hidden Markov Model and then train this model using the Baum-Welch algorithm with only the set of transition states as the input data.
% \newline

% \par
Table~\ref{tbl:model-comp} shows the mean and the standard deviation across 100 runs of the experiment for the different interactions. It can be seen that our proposed approach with the Transition State Clusters on top of the HMM (TSC-HMM) outperforms the baseline HMM. Furthermore the proposed approach introduces a negligible amount of additional computational overhead.
\begin{table}
\centering
\begin{tabular}{|c|c|c|}
\hline
\thead{Action}              & \thead{HMM[cm]} & \thead{TSC-HMM[cm]} \\
\hline
Handshake & $9.4 \pm 9.2$ & $8.7 \pm 8.9$ \\ 
\hline
Rocket Fistbump & $5.2 \pm 4.1$ & $4.0 \pm 4.0$ \\ 
\hline
Parachute Fistbump & $3.8 \pm 5.0$ & $1.7 \pm 2.5$ \\ 
\hline
% Bimanual Handovers & $6.9 \pm 2.9$ & $5.2 \pm 2.3$ \\ 
% \hline
\end{tabular}
\vspace{0.5em}
\caption{This table shows the Mean Squared Error and standard deviation for the different interactions and in centimeters resulting from 100 runs across all trajectories within the test dataset. In each run, the models are trained on a randomly selected batch of 15 demonstrations.}
    \label{tbl:model-comp}
\vspace{-3em}
\end{table}
\section{Conclusion and Future Work}
In this work, we present a framework for segmenting and learning Human-Robot Interactions (HRI) using Hidden Markov Models (HMMs) and Transition State Clustering (TSC). 
% Our Imitation Learning approach directly learns from Human-Human Interaction data, avoiding the need for reinforcement learning or kinesthetic teaching. We emphasize the importance of natural, human-like interaction for widespread robotics adoption, including social interactions and handovers. Hidden Markov Models excel in generalization by segmenting demonstration trajectories, making complex interactions manageable. 
We find that the mismatch in the forward variable calculation between the training and testing scenarios helps identify the observations corresponding to the transition states. Learning a second HMM over these observations leads to an improvement in predicting the robot motions as compared to using a simple HMM, which we demonstrated through different interactive tasks such as handshakes and fistbumps.
% \par
% A major limitation of this work is the absence of testing in both simulation and real-world user studies on a robot. The lack of simulation testing hinders the assessment of the proposed framework's performance under controlled and diverse scenarios. Simulations offer a controlled environment to evaluate adaptability and robustness. Real-world user studies are essential to assess practical utility and user-friendliness, providing valuable insights into system performance and user interactions. Addressing these limitations would significantly enhance the credibility and applicability of the framework.
\newline
However, our approach currently depends on an 'oracle' for action recognition, lacking an automated mechanism for classifying the interaction to be performed. To enhance autonomy, integrating an action recognition method could automate interaction recognition. Additionally, our approach requires manual determination of Hidden Markov Model states, impacting flexibility. An automated solution, possibly using the G-Means algorithm proposed in~\cite{hamerly2003learning}, could improve flexibility, but direct application first requires some adaptation.

\section*{Acknowledgements}
This work was supported by the German Research Foundation (DFG) Emmy Noether Programme (CH 2676/1-1), the German Federal Ministry of Education and Research (BMBF) Projects IKIDA (Grant no.: 01IS20045) and KompAKI (Grant no.: 02L19C150), the EU Projects MANiBOT and ARISE, and the Excellence Program, “The Adaptive Mind”, of the Hessian Ministry of Higher Education, Science, Research and Art.

%%
%% Print the bibliography
%%
\printbibliography

%%
%% If your work has an appendix, this is the place to put it.
\end{document}